\documentclass{article}

\usepackage{PRIMEarxiv}

\usepackage[utf8]{inputenc} 
\usepackage[T1]{fontenc}    
\usepackage{hyperref}       
\usepackage{url}            
\usepackage{booktabs}       
\usepackage{amsfonts}       
\usepackage{nicefrac}       
\usepackage{microtype}      
\usepackage{lipsum}
\usepackage{fancyhdr}       
\usepackage{graphicx}       

\usepackage{tabularx}
\usepackage{CJKutf8}
\usepackage{makecell}

\graphicspath{{media/}}     

\title{CourseGPT-zh: an Educational Large Language Model Based on Knowledge  Distillation Incorporating Prompt Optimization
\thanks{
\textbf{Xing zhang is the corresponding author.}} 
}

\author{
  Zheyan Qu, Lu Yin, Zitong Yu, Wenbo Wang, Xing zhang$^{\ast}$ \\
  Wireless Signal Processing and Network Laboratory, Beijing University of Posts and Telecommunications, Beijing \\
  Department of Computer Science,  University of Aberdeen, UK \\
  zhangx@ieee.org \\
}

\begin{document}
\maketitle

\begin{abstract}
Large language models (LLMs) have demonstrated astonishing capabilities in natural language processing (NLP) tasks, sparking interest in their application to professional domains with higher specialized requirements. However, restricted access to closed-source LLMs via APIs and the difficulty in collecting massive high-quality datasets pose obstacles to the development of large language models in education fields of various courses. Given these challenges, we propose CourseGPT-zh, a course-oriented education LLM that supports customization and low-cost deployment. To address the comprehensiveness and diversity requirements of course-specific corpora, we design a high-quality question-answering corpus distillation framework incorporating prompt optimization, which effectively mines textbook knowledge and enhances its diversity. Moreover, considering the alignment of LLM responses with user needs, a novel method for discrete prompt optimization based on LLM-as-Judge is introduced. During optimization, this framework leverages the LLM's ability to reflect on and exploit error feedback and patterns, allowing for prompts that meet user needs and preferences while saving response length. Lastly, we obtain CourseGPT-zh based on the open-source LLM using parameter-efficient fine-tuning. Experimental results show that our discrete prompt optimization framework effectively improves the response quality of ChatGPT, and CourseGPT-zh exhibits strong professional capabilities in specialized knowledge question-answering, significantly outperforming comparable open-source models.
\end{abstract}


\section{Introduction}
Large language models, such as ChatGPT \cite{chatgpt2023optimizing}, GPT4 \cite{achiam2023gpt}, LLaMA \cite{touvron2023llama}, and ChatGLM \cite{zeng2022glm}, have demonstrated remarkable performance and generalization capabilities across various NLP tasks, significantly expanding the boundaries of language applications. With the increase in model parameters and pretraining corpus size, capabilities such as logical reasoning, instruction following, and In-Context Learning \cite{wei2022emergent},\cite{wei2023larger},\cite{saparov2022language} have emerged. Based on these breakthroughs, the latest LLMs have shown profound understanding and professionalism in various fields, such as virtual assistants, text generation, and code annotation. Utilizing LLMs to disrupt industries has become an inevitable trend, including the field of education\cite{dan2023educhat},\cite{gu2023xiezhi}.

Recently, there has been a desire to leverage the extensive knowledge of large language models to construct domain-specific LLMs in various vertical fields, which require greater expertise and accuracy. To address the issue that general-purpose LLMs cannot meet specific domain requirements, a variety of methods have been proposed. For instance, steering foundation models through role-playing or prompt engineering have been used to tap into the knowledge learned during the pre-training phase, which can unleash their deep-seated expert capabilities \cite{tang2023medagents},\cite{nori2023can}. Other approaches involve pretraining or continual pre-training with domain-specific corpus to incorporate domain-specific knowledge into large language models \cite{dan2023educhat},\cite{tan2023medchatzh},\cite{zhang2023xuanyuan},\cite{le2022bloom}. In addition, to reduce the hallucination during the response generation, retrieval augmentation has also been applied to provide reliable references \cite{dan2023educhat},\cite{cui2023chatlaw}. Based on these approaches, successful implementations such as MedAgents \cite{tang2023medagents}, ChatLaw \cite{cui2023chatlaw}, EduChat \cite{dan2023educhat}, and FinGPT \cite{yang2023fingpt} have demonstrated the potential of LLMs to provide professional responses and insights in various vertical fields, including healthcare, law, finance, and education.

However, constructing domain-specific large language models is still labor-consuming and expensive. To begin with, for closed-source large language models like ChatGPT, the high costs of text generation and fine-tuning services are often prohibitive. As for open-source LLMs, there is a significant gap in parameter size and pre-training corpus compared to closed-source LLMs, resulting in significantly weaker general capabilities such as reasoning, and domain-specific knowledge extraction \cite{gu2023xiezhi},\cite{huang2024c},\cite{zhong2023agieval},\cite{li2023cmmlu}. Faced with complex professional terminology, open-source large language models often fail to meet user requirements for domain knowledge. In this context, it often requires a large amount of in-domain pre-training corpus or expertise datasets to enhance professionalism in vertical fields.

Although various existing works have developed specialized datasets and evaluation criteria for various fields such as philosophy, medicine, and law, as well as for scenarios including network operation and geospatial semantics \cite{huang2024c},\cite{zhong2023agieval},\cite{li2023cmmlu},\cite{guo2023owl},\cite{li2023geoglue}, there is still a considerable demand for manual effort in constructing datasets for courses or privatized scenarios that are not covered by these datasets. This challenge is particularly pronounced when accessible corpora in the field are scarce, making it extremely difficult to construct tens of thousands of specialized instruction data. Furthermore, the majority of models are primarily pre-trained on English corpora, which may lead to a degradation in their performance in other languages \cite{huang2023not},\cite{zhu2023extrapolating}.

In addition to the challenges of constructing specialized corpora, the high cost of inference incurred by open-source large language models cannot be overlooked. Compared to the concise responses provided by humans, the responses generated by large language models, while more comprehensive, also include a significant amount of redundant information, resulting in unnecessary inference overhead. Typically, to further align the responses of large language models with specific preferences, methods such as RLHF (Reinforcement Learning from Human Feedback)\cite{ouyang2022training} are introduced for fine-tuning models. However, this approach still requires a substantial amount of human-labeled preference data. Consequently, promoting alignment between the responses and human preferences, as well as reducing inference costs, is also a key factor in fostering the widespread adoption of open-source large models in specialized vertical domains.

Targeted at these issues, we propose CourseGPT-zh, an open-source education large language model, and design a pipeline for constructing high-quality question-answer pairs through mining textbook knowledge. By utilizing the constructed diverse question-answer pairs, we perform parameter-efficient fine-tuning on the open-source model to mitigate the resource constraints required for deployment. In addition, in the data construction process, we incorporate LLM-as-Judge and utilize discrete prompt optimization to generate optimal prompts, steering ChatGPT to produce high-quality training data aligned with human preferences. Through this method, we ensure high-quality responses while reducing the deployment costs associated with response length.

Our main contributions can be summarized as:
\begin{itemize}
    \item In this paper, we propose CourseGPT-zh, an open-source education large language model, with a pipeline for constructing high-quality and diverse question-answer pairs. Based on textbooks, we guide the model to conduct thorough exploration and questioning of textbooks, extracting knowledge from both closed-source large language models and specialized texts. Additionally, we employ a method inspired by self-instruct to guide the large language models in generating related questions, further enhancing the diversity.
    \item Considering that although large language models can generate comprehensive answers, some content may be redundant or incorrect. Therefore, we employ prompt engineering to guide ChatGPT in generating responses that align with human preferences. To obtain the optimal prompts, we have designed an iterative discrete prompt optimization framework, which incorporates LLM-as-Judge to facilitate automatic evaluation of the quality of responses guided by prompts. Furthermore, the optimized prompt allows the large language model to achieve a balance between the quality of responses and their length, achieving information compression in responses.
    \item A parameter-efficient fine-tuning method of the ChatGLM3 model is conducted based on constructed high-quality question-answering data, resulting in the CourseGPT-zh. Experimental evidence has shown that CourseGPT-zh exhibits improved alignment with human responses, and delivers more concise answers while maintaining a high level of response quality. On various NLP task evaluation metrics, CourseGPT-zh significantly outperforms other open-source large models.
\end{itemize}

\section{Related-work}
\label{sec:headings}

With fierce competition and rapid development, large language models ranging from billions to trillions of parameters have achieved remarkable performance across various NLP tasks after being pre-trained on massive amounts of text. Represented by LLMs such as ChatGPT, GPT4, and GPT4-Turbo, the OpenAI model family has successively reset the benchmarks for NLP tasks, being regarded as one of the greatest inventions in history. Concurrently, a multitude of open-source large language models, including llama-2-13b, ChatGLM3-6b, and Mistral-8x7B-MoE\cite{jiang2024mixtral}, have also shown astonishing improvements, even surpassing the level of ChatGPT on some dimensions. More importantly, they can be deployed on a single to several GPUs and can be flexibly customized through fine-tuning.

\subsection{Domain-specific LLMs}
Although general-purpose large language models have achieved exceptional performance on generic NLP tasks, they often fall short in vertical domains that necessitate extensive specialized knowledge and high accuracy requirements. The performance of zero-shot large language models in these domains is typically inadequate, thereby granting domain-specific LLMs significant attention. Closed-source large language models, while exhibiting superior performance across various capabilities, present challenges for continual pre-training and fine-tuning with private corpora. Therefore, the construction of domain-specific models based on closed-source LLMs frequently leverages role-playing or collaboration abilities to extract knowledge in the specialized field during the pre-training phase. In contrast, open-source LLMs can be further pre-trained or fine-tuned with extensive high-quality domain-specific data, and they have achieved multiple successful applications in fields such as medicine, law, education, finance, etc.

HuatuoGPT \cite{zhang2023huatuogpt} employs a mixed dataset comprising distilled data from ChatGPT and real-world data provided by physicians' medical advice to fine-tune an open-source model. Furthermore, it aligns the model's response with human preferences through RLAIF (Reinforcement Learning from Artificial Intelligence Feedback). By learning from the response styles of real-world doctor-patient interactions, the fine-tuned model can engage with users in a human-like manner and significantly surpasses other models at a similar level across various metrics. MedChatZH \cite{tan2023medchatzh} has developed a dialogue model specifically designed for Traditional Chinese Medicine, incorporating extensive Chinese medical literature for continual pre-training. After fine-tuning millions of question-answer data from the Internet and various Chinese hospitals, the model achieves state-of-the-art performance in the field of Chinese medicine. ChatLaw \cite{cui2023chatlaw}, targeting the legal domain, not only provides professional responses concerning legal knowledge but also acquires problem-solving abilities through training on multiple-choice question data. Furthermore, it employs a method combining vector database retrieval with keyword search, effectively reducing the hallucination in responses. EduChat \cite{dan2023educhat} offers a range of functionalities, including open-ended question answering, paper assessment, and Socratic teaching, enhancing various skills through fine-tuning and the integration of tools. The model gains interdisciplinary knowledge through continual pre-training and strengthens its question-answering and instruction-following capabilities with large-scale instruction and open-domain dialogue datasets. FinGPT \cite{yang2023fingpt} adopts a data-centric approach, focusing on automated data management pipelines and lightweight adaptive technologies, establishing a comprehensive framework from data processing to feature engineering and application, while also enhancing the transparency of the overall framework. One of its strengths lies in its ability to integrate seamlessly with both open-source and closed-source large language models without the need for further training.

\subsection{Discrete prompt engineering}

Prompt engineering aims to guide large language models to fully leverage their potential through the meticulous design of prompts. Extensive research has demonstrated that well-crafted prompts can significantly enhance the ability of large language models to improve their performance across various NLP tasks \cite{liu2023gpt},\cite{wei2022chain}. Prompt engineering encompasses continuous prompt learning and discrete prompt optimization. Continuous prompt learning aims to adapt large language models to various tasks by incorporating learnable parameters within the prompts \cite{li2021prefix}, \cite{liu2021p}. However, continuous prompt learning typically requires access to the gradient vectors of the LLMs, which restricts its application in closed-source models that are accessed only through APIs. For discrete prompts, traditional methods often rely on meticulous manual design, which not only demands considerable human effort but also may not necessarily maximize the model's performance. Consequently, numerous methods for automatically generating optimal discrete prompts have been explored, leveraging the large model itself as an optimizer to autonomously enhance its performance in NLP tasks.

Recently, several leading automated discrete prompt optimization frameworks have been proposed. EVOPROMPT\cite{guo2023connecting} draws on the principles of evolutionary algorithms (EAs) to iteratively guide LLMs to generate new prompts through evolutionary operators. It does not require any gradient information from LLMs and can achieve a balance between exploration and exploitation. Experiments on nine datasets have shown that optimized prompts can significantly improve task performance. APE\cite{zhou2022large}, inspired by program synthesis, represents discrete prompting optimization as a black-box optimization problem. It treats instructions as "programs" and optimizes them by searching through the candidate instruction pool proposed by LLMs. Furthermore, it employs an iterative Monte Carlo search to further enhance prompt performance. OPRO\cite{yang2023large} utilizes LLMs to generate new candidate prompts from previously generated results and their scores and then evaluates the new candidate prompts for the next iteration. PROMPTAGENT\cite{wang2023promptagent} approaches it as a strategic planning problem, using Monte Carlo tree search to achieve a balance between exploration and exploitation. Unlike other discrete prompt optimization frameworks, it also leverages learning capabilities based on error summarization of large language models, introducing expert-level domain knowledge and guidance based on reflection. The optimal prompts obtained from these prompting optimization frameworks have achieved results significantly better than manually crafted prompts on various NLP tasks, including GSM8K and Big-Bench Hard tasks.

\begin{figure}
  \centering
    \includegraphics[width=120mm]{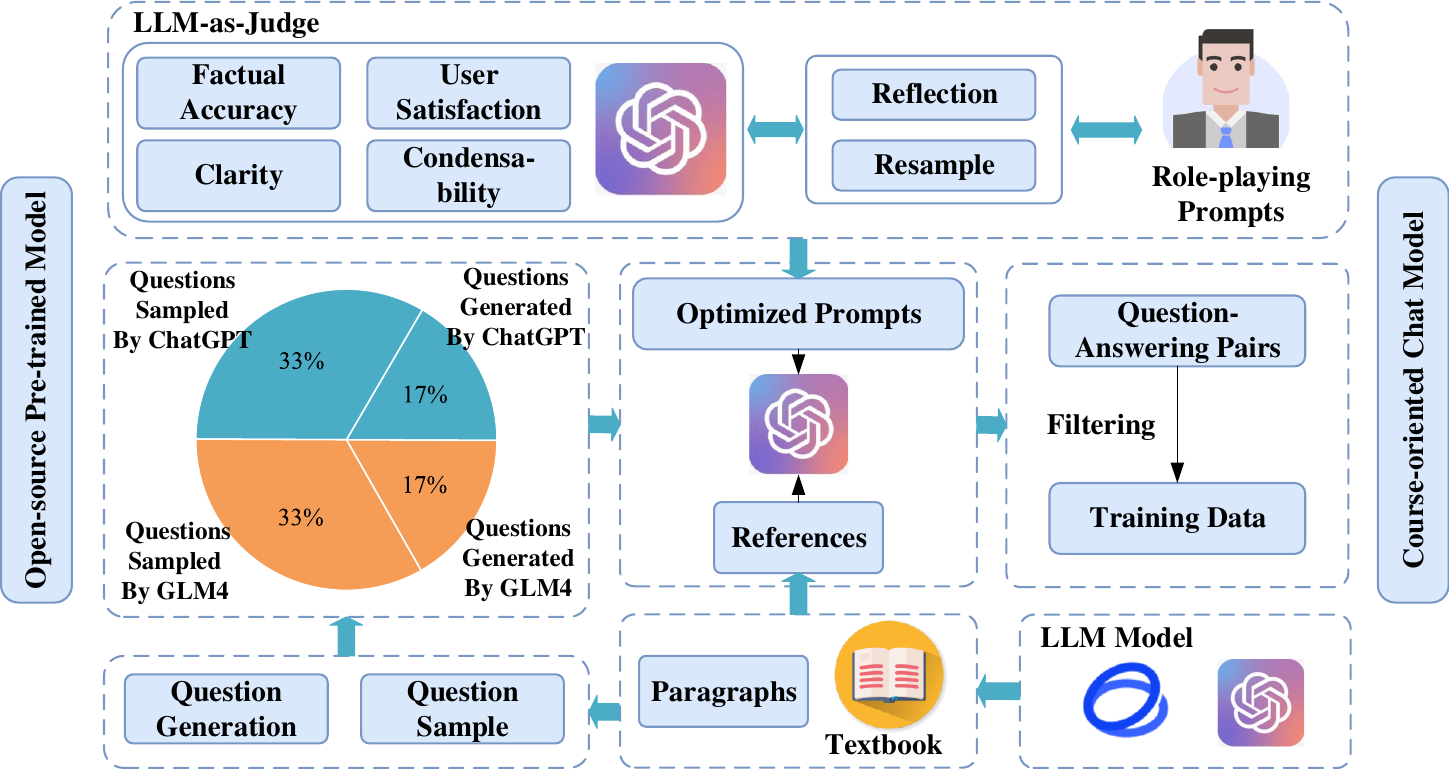}
  \caption{CourseGPT-zh Framework}
  \label{fig:fig1}
\end{figure}

\section{Data Construction}
\label{sec:headings}

Large language models often require at least tens of thousands of high-quality instruction-tuning data to demonstrate satisfactory performance; however, the collection and processing of such data can be prohibitively labor-intensive. Unlike fields such as medicine, which benefit from a wealth of open-source question-answering datasets gathered from the internet and hospital medical databases, amassing large volumes of high-quality question-answering datasets poses significant challenges. In light of these barriers to data construction and the demand for low-cost model development, we propose a pipeline based on knowledge distillation from ChatGPT and GLM-4, which ensures the comprehensiveness and diversity of questions, as well as the professionalism and alignment of the distilled responses. The entire process consists of two components: question construction and response generation.

\subsection{Question Generation}

To ensure that the fine-tuned model can provide professional answers to users' questions about various knowledge both inside and outside the textbooks, the comprehensiveness and diversity of the questions are of great importance. Merely relying on human-generated questions as a seed pool and using methods like self-instruct \cite{wang2022self} to generate question data may not cover all knowledge points comprehensively, especially for the understanding of various professional terms in specialized fields. In addition, the diversity of questioning methods for the same set of knowledge points also needs to be addressed. Diverse questioning and answering of the same knowledge can guide the model to learn the internal relationships of knowledge more effectively.

In response to these challenges, we have developed a diversified question generation pipeline, as depicted in the orange section of Figure \ref{Data Construction Framework}. Initially, knowledge extraction and questioning are based on textbook paragraphs to ensure the comprehensiveness of the questions. The textbook is divided into paragraphs, which are sequentially input into a large language model to guide the generation of a list of questions targeting specific knowledge points. Concurrently, during the process, 6 questions from a seed pool comprising 50 carefully chosen human-written questions, along with two generated questions are randomly selected as in-context question examples. This approach steers the model towards generating diverse questions for a group of knowledge points, such as interpretive questions, pros and cons comparison questions, and comparative questions. Lastly, in terms of question quantity, we force the large language model to generate an excess number of questions for the limited length of paragraphs. This serves to distill the knowledge from the large model and enhance the diversity of the questions.

However, the question lists constructed based on textbook paragraphs are still limited in form and content by the textbook. To further enhance the diversity, inspired by self-instruct, we employ an iterative approach to distill new questions from a large model. Specifically, a set of questions generated in the previous round is selected as content examples, guiding the large language model to sample and generate new content-related questions. At the same time, 3 questions are randomly selected from the seed pool as style examples to enrich the diversity of question forms. After generating a new set of questions, it is then used as content examples for the next iteration. In this way, we can reduce the influence of referencing textbook paragraphs, and a large number of new questions can be obtained through distillation.

Finally, we find that different large language models exhibit significant variation in the style and distribution of the questions they generate. Consequently, we employ both ChatGPT and GLM-4 for concurrent question generation and sampling, and subsequently deduplicated the final question list. As CourseGPT-zh is an adaptable course-specific large language model, we selected Communication Principles as the focus of our experiment, which encompasses specialized domains such as signal modulation, quantization, and coding, requiring the model to provide professional and precise responses. Acknowledging the limitations of open-source models in mathematical derivation capabilities such as integration, our focus is directed towards the learning of conceptual knowledge in this field. Utilizing the textbook on Communication Principles, we separately generate approximately 10k questions using ChatGPT and GLM-4 and conduct two rounds of iterative sampling based on these questions, resulting in about 20k sampled questions.

\subsection{Answer Generation}

\begin{figure}
  \centering
    \includegraphics[width=120mm]{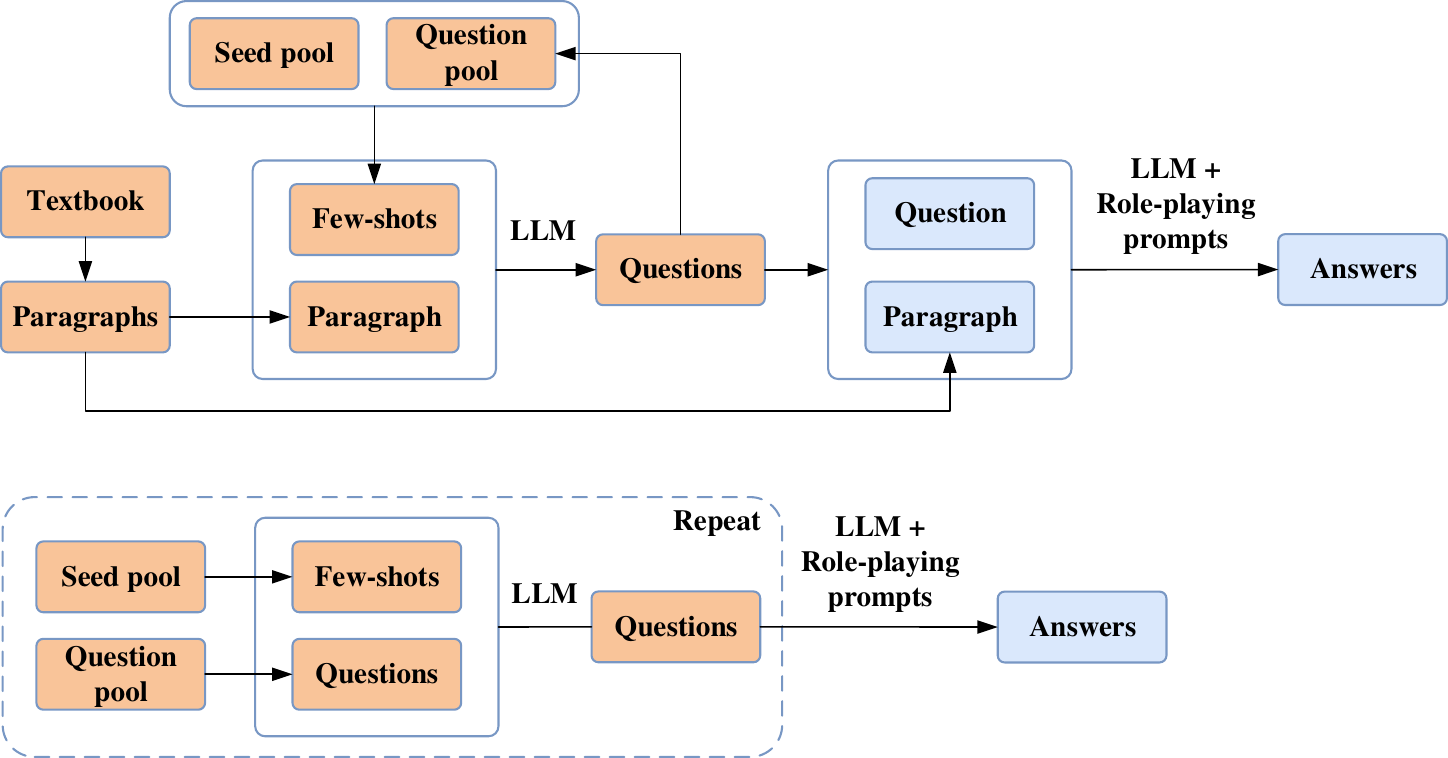}
  \caption{Data Construction Framework}
  \label{Data Construction Framework}
\end{figure}

After obtaining a diverse list of questions, the next challenge is to utilize large language models to generate professional answers that meet users' preferences and needs. Generally, LLMs tend to produce comprehensive and well-structured answers that are reader-friendly. However, these lengthy responses not only significantly differ from the style of human replies but also substantially increase response latency and reasoning. Therefore, it is necessary to align the response style of LLMs with that of humans and enhance the accuracy and professionalism of the responses. As shown in the blue part of the Figure \ref{Data Construction Framework}, we leverage role-playing prompts to guide the large language model in generating responses that meet the requirements. Taking advantage of role-playing capabilities, the large language model can focus on professional fields and generate more reliable responses. For the same group of tasks, different prompts can lead to significant differences in model performance, which is also the case in the question-answering field. It is worth noting that for questions derived from textbook paragraphs, we refer to the original text to improve the accuracy of the answers. The optimization of prompts will be introduced in the next section.

\section{Discrete Prompt Optimization}
\label{sec:headings}

Research indicates that the design of prompts significantly affects the performance of large language models on NLP tasks \cite{liu2023gpt},\cite{wei2022chain}. Prior studies have emphasized the design of prompts at various stages, yet the prompts deemed optimal by humans may not necessarily elicit the desired performance. Therefore, optimizing the prompts is essential, particularly during the process of knowledge distillation in specialized domains, which can prompt the large language models to generate more professional responses. In our discrete prompt optimization framework, we further incorporate LLM-as-Judge to evaluate the alignment between the responses guided by prompts and human responses. By training on the distillation data guided by optimal prompts, the fine-tuned large language model obviates the need for further optimization using Reinforcement Learning from Human Feedback (RLHF).

\subsection{LLM-as-Judge}

Alignment is a crucial step that ensures the model's performance aligns with user intentions and human preferences. Generally, Reinforced Learning from AI Feedback (RLAIF) is extensively applied in alignment tasks. However, reward models can be challenging to train and are susceptible to the influence of human errors in the training data. Moreover, in contrast to the approach of first training with mixed data and then aligning, we aim to maintain alignment between LLM responses and human responses starting from the construction of the training data. This method avoids potential issues such as convergence difficulties and reward hacking in reinforcement learning models \cite{rame2024warm},\cite{chen2024odin}. Furthermore, recent research has started to introduce large models, such as ChatGPT, as judges for alignment evaluation. By comparing LLM-generated responses with human responses across multiple dimensions, LLMs have demonstrated a high degree of consistency with human judgments \cite{zheng2024judging},\cite{liu2023alignbench}. More importantly, it requires only a subset of validation samples to effectively reflect the overall situation and can quickly adjust the evaluation dimensions according to user needs, significantly reducing costs.

In this section, we adopt the prompt design of the judge in AlignBench \cite{liu2023alignbench}, taking into account Factual Accuracy, User Satisfaction, Clarity, and Condensability. We separately score each evaluation dimension using CoT to obtain the final score, thereby enhancing the transparency and credibility of the evaluation process and improving alignment performance. Factual Accuracy considers the accuracy of information in the LLM response and whether it aligns with the facts presented in human responses. User Satisfaction evaluates whether the LLM response adequately and appropriately meets the user's query needs in comparison to human responses. Clarity aims to assess whether the LLM response is concise and clear in comparison to human responses, enhancing user readability. Lastly, Condensability evaluates whether the LLM response is succinct and refined, considering the potential redundancy in the LLM response. Furthermore, by collecting evaluation results, we can understand the strengths and weaknesses of the current prompts across various dimensions, providing instructive feedback for further prompt improvement. Based on this feedback, we can further refine the prompts using reflection \cite{shinn2023reflexion}.

\subsection{Discrete Prompt Optimization Framework}

The application of knowledge distillation using large language models such as ChatGPT has been demonstrated to effectively train student models, thereby enhancing their performance across various NLP tasks \cite{zhang2023huatuogpt}, \cite{hsieh2023distilling}. However, the quality of responses generated by ChatGPT heavily depends on the design of the prompts, which has been overlooked in previous work within specialized domains. Different prompts significantly affect the style, accuracy, and professionalism of the responses. Considering the refined language and precise factual responses typical of human replies, we aim to fine-tune the model using distillation data guided by optimal prompts. This will enable the fine-tuned model to enhance the information density of its responses while meeting user requirements, thereby avoiding unnecessary output and improving generation speed.

In this case, we have developed a discrete prompt optimization framework based on reflection \cite{shinn2023reflexion} and LLM-as-Judge. In this framework, we task LLM-as-Judge with scoring candidate prompts on a randomly sampled set of 50 human-drafted question-answer pairs to assess the quality. The evaluation feedback provided by LLM-as-Judge offers a clear direction for the improvement of the candidate prompts. Consequently, we integrate the feedback with Reflection to guide the learning process of the LLM, thereby enhancing the quality of the prompts. It is noteworthy that LLM-as-Judge initially provides separate evaluations for each dimension and concludes with a comprehensive evaluation and scoring. To reduce the overhead associated with the reflection operation, we randomly collect 5 comprehensive evaluation results as feedback for each candidate prompt, which serve as the basis for subsequent improvements.

However, in this process, we found that although utilizing evaluation results to make improvements could enhance the performance of the prompts and make them more aligned with the evaluator's criteria, it simultaneously tends to generate overly lengthy prompts and encourages ChatGPT to produce longer responses to achieve higher scores. This might lead to deviations from human-like response styles and increase the cost of inference. To address this issue, we took the following measures: Firstly, in addition to the LLM scores, we introduced a length penalty factor by referencing the length of human responses to more reasonably balance the comprehensiveness and length of the responses. Secondly, during the prompt optimization process, besides improving the prompts with feedback, we incorporated a Resample module. Specifically, after calculating the scores of candidate prompts using LLM-as-Judge, we select the top five highest-scoring prompts sorted in descending order. These sorted prompt-score pairs are then used as input to guide the large language model to automatically discover the key information within the prompts and the patterns related to the scores, thereby resampling an equivalent number of prompts. This approach allows for the retention of key information within the prompts while also achieving their conciseness.

\begin{figure}
  \centering
    \includegraphics[width=120mm]{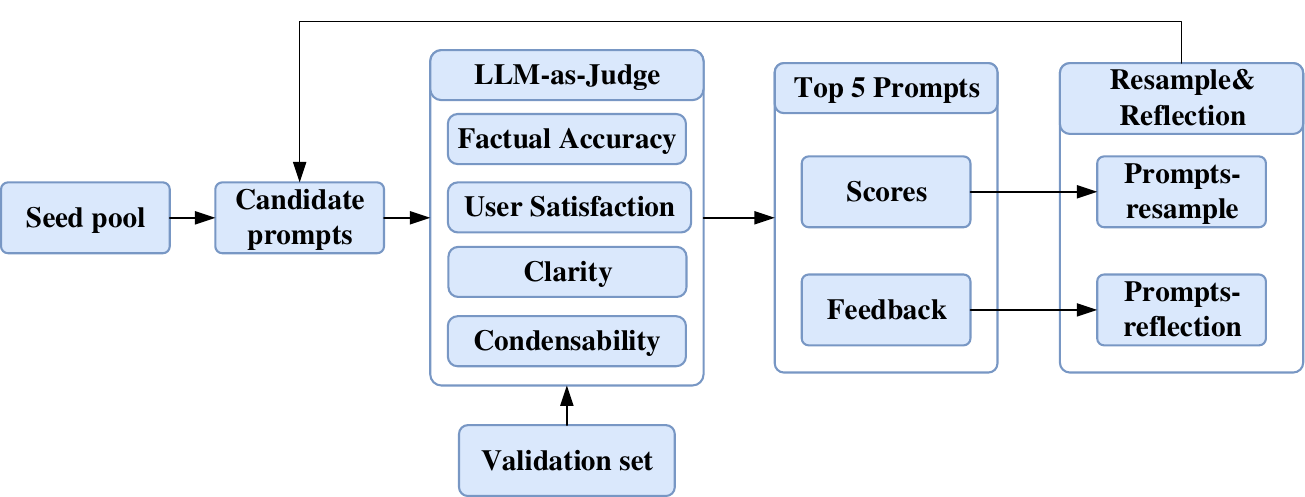}
  \caption{Discrete Prompt Optimization Framework}
  \label{Discrete Prompt Optimization Framework}
\end{figure}

As illustrated in Figure \ref{Discrete Prompt Optimization Framework}, the proposed optimization framework for discrete prompts is presented. Initially, a set of ten manually crafted prompts are employed as the preliminary candidates, and their respective scores and evaluation outcomes are obtained using LLM-as-Judge. The significance of these initial prompts lies in their capacity to inject the preliminary requirements into the optimization framework for further refinement. It is noteworthy that role-playing is incorporated into the initial prompts to guide the large model to concentrate on professional domains, thereby generating more specialized responses.


In terms of scoring, the calculation method is as described by Equation \ref{eq}, where \textit{$s_{i}^{LLM}$} represents the comprehensive score given by LLM-as-Judge, \textit{$l^{res}$} denotes the length of the response generated by the LLM, and \textit{$l^{ref}$} indicates the length of the reference human response. If the LLM-generated response is shorter than the reference length, no penalty is imposed, and the evaluation is solely based on the LLM-as-Judge's assessment across various dimensions. Conversely, if the LLM-generated response exceeds the reference length, a penalty is applied to the additional length, with the degree of penalty determined by the parameter \textit{$\alpha$}.

\begin{equation}
\textit{$s_i$} = \left\{ \begin{array}{ll}  
s_{i}^{LLM} & {l^{res}\le l^{ref}}\\
s_{i}^{LLM}-\alpha(\frac{l^{res}}{l^{ref}}-1) & {l^{res}>l^{ref}}
\end{array} \right.
\label{eq}
\end{equation}

After obtaining the scores and feedback for each candidate prompt, we iteratively generated the next generation of candidate prompts by combining reflection and resampling modules. Initially, we selected the top 5 prompts with the highest scores for subsequent operations. The reflection method effectively utilizes the evaluation results from LLM-as-Judge to refine the prompts, ensuring that the generated prompts better meet the requirements of multiple dimensions. Furthermore, by incorporating the resampling module, the LLM can identify patterns and key components, sampling a new generation of prompts. During optimization, the top 5 prompts might include candidates generated from the two aforementioned modules in the previous iteration. In such cases, the resampling module can leverage the key information from the prompts optimized by both modules.

In the experiment, we set the parameter \textit{$\alpha$} to 0.5 and conducted three iterations, and we selected the prompt term with the highest score on the validation set as the optimal prompts.

\section{Experiment and Analysis}
\label{sec:headings}

\subsection{Model Training}

We conducted fine-tuning based on ChatGLM3-6B as the foundational architecture. ChatGLM3-6B is a pre-trained large language model tailored for Chinese, which possesses excellent features such as conversational fluency and low deployment threshold. In the instruction fine-tuning, we employed the LoRA strategy \cite{hu2021lora}, with the rank textbf{\textit{k}} set to 64 and textbf{\textit{alpha}} set to 128. Additionally, the learning rate, batch size, and maximum context length were set to 1e-4, 32, and 2048, respectively. Regarding training facilities, we distributed the model across 8 A40 GPUs using Pytorch to accelerate the training process.

\subsection{Benchmarks}

We constructed a test dataset consisting of 200 QA pairs derived from examination papers and web pages in the field of communication principles, encompassing various chapters within this domain to evaluate the model's proficiency in various knowledge. In our tests, we selected several open-source and closed-source models that support Chinese, including Qianfan-llama2-7b-chinese, Qianfan-llama2-13b-chinese, chatglm3-6B, ERNIE-Bot-turbo \cite{sun2021ernie}, and ChatGPT, for comparative purposes.

Among them, Qianfan-llama2-7b-chinese and Qianfan-llama2-13b-chinese are versions based on Llama-2 that have undergone enhanced pre-training with large-scale Chinese-English corpora and fine-tuning for instruction following, which has improved their performance in Chinese-English question-answering. ChatGLM3-6B is a bilingual dialogue language model released by Zhipu AI and the KEG Laboratory of Tsinghua University, demonstrating optimal performance among models below 10B parameters in semantics, mathematics, reasoning, and coding. ERNIE-Bot-turbo, developed by Baidu, encompasses a vast amount of Chinese data, exhibiting impressive capabilities in Chinese question-answering and Chinese content generation. Lastly, ChatGPT is recognized as one of the most advanced models. To verify the effectiveness of prompt optimization, we also conducted a comparative analysis of the response quality of GPT-3.5-turbo under the guidance of optimal prompts.

\subsection{Evaluation Metrics}

We employed the BLEU\cite{papineni2002bleu}, GLEU\cite{mutton2007gleu}, and ROUGE\cite{lin2004rouge} metrics to evaluate the similarity between the responses generated by LMM and the reference human responses. BLEU is an accuracy-based evaluation method that assesses the similarity between LLM responses and reference responses through the overlap precision of n-grams. GLEU further takes into account factors such as lexical overlap and order, reflecting the fluency and naturalness of sentences. Unlike BLEU, ROUGE primarily focuses on the recall of n-grams, based on the comprehensiveness and coverage of the LLM responses. Finally, ROUGE-L is based on the calculation of the longest common subsequence of matches.

However, these metrics are solely based on n-grams matching and cannot evaluate the alignment of semantics between LLM responses and reference human responses. Therefore, we also incorporated LLM-as-Judge to assess the quality of responses in terms of factual accuracy, user engagement, clarity, and conciseness. Additionally, we recorded the LLM-as-Judge scores, length penalty scores, and comprehensive scores to accurately and comprehensively reflect the quality of the responses.

\subsection{CourseGPT-zh}

All models were scored on traditional Natural Language Processing (NLP) metrics as detailed in Table \ref{table1}. For each comparative model, we utilized the officially provided APIs or checkpoint models to obtain responses on the test set. As depicted in the table, closed-source models such as ERNIE-Bot-turbo and ChatGPT exhibit significant advantages over open-source models due to their larger parameter counts and the benefits of pre-trained corpora. Additionally, compared to the performance of ChatGPT without the use of additional prompts, the application of optimal prompts led to effective improvements across all metrics for ChatGPT, particularly enhancing the fluency of responses by approximately 17$\%$. This validates the efficacy of the discrete prompt optimization framework we proposed. Surprisingly, the fine-tuned ChatGLM3 surpassed ChatGPT on all metrics, especially achieving a higher recall rate in the ROUGE metric. This may be attributed to its training on the data distilled under the guidance of optimal prompts.

\begin{table}[h]
\small
\centering
\begin{tabular}{lllllllll}
\hline
Model                      & BLEU-1         & BLEU-2         & BLEU-3         & BLEU-4         & GLEU           & ROUGE-1        & ROUGE-2        & ROUGE-L        \\ \hline
Qianfan-llama2-7b-chinese  & 0.157          & 0.073          & 0.037          & 0.018          & 0.056          & 0.254          & 0.059          & 0.193          \\
Qianfan-llama2-13b-chinese & 0.176          & 0.081          & 0.040          & 0.021          & 0.060          & 0.258          & 0.062          & 0.190          \\
ChatGLM3-6B                & 0.166          & 0.075          & 0.039          & 0.020          & 0.056          & 0.245          & 0.053          & 0.180          \\
ERNIE-Bot-turbo            & 0.180          & 0.083          & 0.041          & 0.021          & 0.060          & 0.247          & 0.057          & 0.178          \\
ChatGPT                    & 0.229          & 0.108          & 0.055          & 0.028          & 0.077          & 0.274          & 0.068          & 0.202          \\
\hline
ChatGPT-prompt             & \textbf{0.254} & 0.117          & 0.059          & 0.030          & \textbf{0.090} & \textbf{0.300} & 0.072          & \textbf{0.222} \\
CourseGPT-zh              & 0.253          & \textbf{0.120} & \textbf{0.063} & \textbf{0.033} & 0.088          & 0.297          & \textbf{0.076} & 0.218          \\ \hline
\end{tabular}
\caption{Benchmark on QA dataset.(\textbf{zero-shot})}
\label{table1}
\end{table}

However, these traditional metrics have clear limitations as they only focus on the calculation of n-gram-related accuracy or recall rates, lacking the ability to evaluate complex semantic information. Therefore, we continued to use LLM-as-Judge to evaluate the response quality of each model, as shown in Table \ref{table2}. Firstly, by comparing the results of the open-source models, it can be observed that in the field of domain-specific question-answering, simply increasing the number of model parameters does not result in significant progress. Instead, training with specialized and high-quality corpora is necessary. The superior performance of ChatGLM3-6b among open-source models may be related to its broader Chinese pretraining corpus. Furthermore, by comparing the results of open-source and closed-source models, it can be seen that although both tend to generate long responses to meet user needs, there are significant differences in response quality. The model with the highest comprehensive score is ChatGPT, while ERNIE-Bot-turbo has the highest response quality, albeit with excessively long responses.

In comparison to the responses generated by ChatGPT under the guidance of optimal prompts, the reduction in length penalty is more than threefold, while achieving nearly identical response quality. This indicates that under the guidance of optimal prompts, the density of effective information in its responses has significantly increased, taking into account multiple dimensions such as accuracy and responsiveness to user needs. Furthermore, the response quality of the fine-tuned ChatGLM3 model significantly surpasses that of various open-source models, and due to its refined responses, it has obtained the second-highest overall score. Compared to the ChatGLM3-6B model without fine-tuning, it achieves better response quality while reducing the length penalty by 63$\%$. This demonstrates that fine-tuning with specialized distillation corpora in a specific style can significantly enhance the response quality of open-source models.

In comparison to the vanilla ChatGPT, the ChatGPT guided by optimal prompts exhibited a reduction in length penalty by more than threefold, while maintaining nearly equivalent response quality. This demonstrates that, under the guidance of optimal prompts, there is a significant enhancement in the density of effective information, satisfying multidimensional requirements such as accuracy and responsiveness to user needs. Furthermore, the fine-tuned ChatGLM3-6B model significantly outperformed various open-source models in terms of response quality, and it achieved the second-highest comprehensive score due to its refined answers. This indicates that fine-tuning with specialized distillation corpora of a specific style can markedly improve the response quality of open-source models.

It is worth noting that the LLM-as-Judge is tested based on ChatGPT. With the evolvement of ChatGPT, the evaluation scores might change in the future, so it is necessary to pay attention to the relative scores.

\begin{table}[h]

\centering
\begin{tabular}{llll}
\hline
Model                      & Comprehensive Score & LLM-as-Judge  & Length Penalty \\ \hline
Qianfan-llama2-7b-chinese  & 4.64                & 5.54          & 0.90           \\
Qianfan-llama2-13b-chinese & 4.92                & 5.84          & 0.92  \\
ChatGLM3-6B                & 5.28                & 6.21          & 0.93           \\
ERNIE-Bot-turbo            & 5.72                & \textbf{6.75} & 1.03           \\
ChatGPT                    & 6.14                & 6.69          & 0.55           \\
\hline
ChatGPT-prompt             & \textbf{6.48}       & 6.65          & \textbf{0.17}  \\
CourseGPT-zh              & 6.21                & 6.55 & 0.34  \\ \hline
\end{tabular}

\caption{The model scores on 200 single-turn questions, using LLM-as-Judge (ChatGPT)}
\label{table2}
\end{table}

\subsection{Discrete prompt optimization framework}

\begin{table*}[!htbp]
	\centering
	\begin{tabularx}{\textwidth}{p{0.5\textwidth}p{0.15\textwidth}p{0.15\textwidth}p{0.15\textwidth}}
 
		\toprule
		\textbf{Prompts} &  \textbf{Comprehensive score} & \textbf{LLM-as-Judge} & \textbf{Length \newline Penalty}  \\
		\midrule
		\begin{CJK*}{UTF8}{gbsn}你是一位通信工程领域的专家，你以简洁明了的方式提供准确无误的回答。你的回答简洁明了、准确无误，避免冗长和繁琐。\newline As an expert in the field of telecommunications engineering, you provide concise and unambiguous responses that are error-free. Your answers are succinct and precise, devoid of superfluity and complexity.\end{CJK*}
		& 6.48 
		& 6.65
		& 0.17
		\\
		\midrule
		\begin{CJK*}{UTF8}{gbsn}你是一位通信工程领域的教授，专注于通信原理领域。以深厚的知识和清晰的表达著称。你的回答结构化、简明扼要、易于理解、准确无误、全面清晰。你提供准确信息，确保回答满足用户需求。\newline As a professor in the field of telecommunications engineering, specializing in communication principles, you are renowned for your profound knowledge and clear articulation. Your responses are structured, concise, and easy to understand, while also being accurate, comprehensive, and clear. You provide accurate information to ensure that your answers meet the user's needs.\end{CJK*}
		& 6.37
		& 7.23
		& 0.85
		\\
        \midrule
		\begin{CJK*}{UTF8}{gbsn}请用简洁明了的语言回答以下问题，确保回答准确无误、全面清晰。请核实事实，满足用户需求，并尽量使用通俗易懂的语言，简化句子结构，提高回答的凝炼性。\newline Please answer the following questions with concise and clear language, ensuring that the answers are accurate, comprehensive, and lucid. Verify the facts, meet the user's needs, and strive to use easily understandable language. Simplify the sentence structure to enhance the conciseness of the responses.\end{CJK*}
		& 5.98
		& 6.17
		& 0.19
		\\
		\bottomrule
	\end{tabularx}%
	
	\caption{Sample prompts from discrete prompt optimization.}
 \label{table3}%
\end{table*}%

As shown in Table \ref{table3}, three types of prompts were collected during the optimization process. The first prompt with the highest comprehensive score was selected as the optimal prompts for subsequent data construction. This prompt utilized role-playing and extracted the most critical semantic components from the optimization experience, achieving a balance between answer quality and length. The second prompt was generated through the reflection module, which effectively integrated the experience of feedback. However, this prompt was overly comprehensive, resulting in the best answer quality but at the cost of excessively long responses. Compared to the non-optimized prompt, its length penalty increased by 0.3. The last prompt, although capable of generating brief responses, failed to guide the generation of high-quality content. This demonstrates that under nearly the same response length, the design of the prompt has a significant impact on the quality of the response. Additionally, during the prompt optimization process, the quality of the prompts did not consistently improve but experienced certain fluctuations.

\subsection{Case study}

\textbf{\emph{Question: What are the impacts of inter-symbol interference?}}

As shown in Table \ref{table4}, we present case studies of responses from different models to the same question. Firstly, ChatGPT guided by optimal prompts provided a refined response and was able to correctly answer the question, covering aspects such as bit error rate and communication quality. However, ChatGLM3-6B explained the concept of inter-symbol interference but failed to correctly respond to the question. Finally, CourseGPT-zh, with a similarly refined style, provided explanations regarding bit error rate, system capacity, and modulation performance, correctly addressing the user's inquiry.

\begin{table*}[!htbp]
        \footnotesize
	\centering
	\begin{tabularx}{\textwidth}{p{0.15\textwidth}p{0.8\textwidth}}
 
		\toprule
		\textbf{Model/References} &  \textbf{Response}  \\
		\midrule
        References
		& \begin{CJK*}{UTF8}{gbsn}1.错误解码:码间串扰可能导致接收端错误地解码接收到的信号……导致数据传输错误或降低解调的准确性。\newline 2.误码率增加:码间串扰增加了误码率……导致信号质量下降。\newline 3.降低系统容量:码间串扰眼制了数字通信系统的容量……系统无法传输高速数据或提供更高的数据传输速率。\newline 4.降低信号质量:码间串扰引入了额外的噪声和失真……影响通信的可靠性和性能。\newline 1. Error Decoding: Inter-symbol interference may lead to the receiver incorrectly decoding the received signal... resulting in data transmission errors or reduced demodulation accuracy. \newline 2. Increased Bit Error Rate: Inter-symbol interference increases the bit error rate... leading to a decrease in signal quality. \newline 3. Reduced System Capacity: Inter-symbol interference constrains the capacity of digital communication systems... the system is unable to transmit high-speed data or provide higher data transmission rates. \newline 4. Degraded Signal Quality: Inter-symbol interference introduces additional noise and distortion... affecting the reliability and performance of communication.\end{CJK*}
		\\
        \midrule
        ChatGPT-prompt
		& \begin{CJK*}{UTF8}{gbsn}码间串扰是指不同码片序列之间相互干扰的现象，在通信系统中会产生以下影响:\newline
1.码误差:码间串扰会号致接收端无法正确解码接收到的信号……降低通信系统的性能。\newline
2.信号衰减:码间串扰会引起信号的衰减……导致通信质量下降。\newline
3.互调干扰:码间串扰还可能引起互调干扰……影响系统的动态范围和信号的有效传输。\newline
4.报文错误率增加:码间串扰会增加报文的错误率……码间串扰对通信系统的性能和质量都会产生不利的影响。\newline
Inter-symbol interference refers to the phenomenon of interference between different chip sequences, which has the following effects in communication systems:\newline
1. Code Error: Inter-symbol interference can cause the receiver to be unable to decode the received signal correctly ... thus degrading the performance of the communication system.\newline
2. Signal Attenuation: Inter-symbol interference can cause signal attenuation ... leading to a decrease in communication quality.\newline
3. Intermodulation Interference: Inter-symbol interference may also cause intermodulation interference ... affecting the dynamic range of the system and the effective transmission of signals.\newline
4. Increased Message Error Rate: Inter-symbol interference can increase the error rate of messages ... inter-symbol interference has an adverse effect on the performance and quality of communication systems.
\end{CJK*}
		\\
		\midrule
        ChatGLM3-6B
		& \begin{CJK*}{UTF8}{gbsn}码间串扰是指两个或多个信号在同一信道中传输时,由于信号之间的相互作用而产生的相互干扰……码间串扰的影响因素包括:\newline
1.信道条件: ……当信道条件较差时, 码间串扰会更严重。\newline
2.信号特征: ……信号的幅度和相位特征可能会导致码间串扰的产生。\newline
3.信道编码:……也能够通过解码恢复出原始信号。\newline
4.信号干扰:……可以采用一些技术, 例如信道编码、自适应滤波器、多址技术等。\newline
Inter-symbol interference refers to the mutual interference that occurs between two or more signals transmitted over the same channel due to their interactions...The factors influencing inter-symbol interference include:\newline
1. Channel conditions: ... inter-symbol interference is more severe when channel conditions are poor.\newline
2. Signal characteristics: ... the amplitude and phase characteristics of the signal may lead to the generation of inter-symbol interference.\newline
3. Channel coding: ... and can also recover the original signal through decoding.\newline
4. Signal interference:... some techniques, such as channel coding, adaptive filters, and multiple access technologies, can be employed to address this issue.
\end{CJK*}
		\\

        \midrule
        CourseGPT-zh
		& \begin{CJK*}{UTF8}{gbsn}码间串扰会产生以下影响:\newline
1.增加误码率:码间串扰会引入额外的噪声……从而增加误码率。\newline
2.降低系统容量:码间串扰会限制不同信号之间的距离……从而降低系统的容量。\newline
3.影响调制性能:码间串扰会改交信号的频谐特性……影响系统的传输质量。\newline
4.增加信号功率:为了抵消码间串扰引起的误码率增加……增加了功耗和成本。\newline
因此……需要采取合适的调制方案、均衡技术和干扰抑制方法来提高系统的性能。\newline
Inter-symbol interference (ISI) results in the following effects:\newline
1. Increased bit error rate: ISI introduces additional noise ..., thereby increasing the bit error rate.\newline
2. Reduced system capacity: ISI constrains the distance between different signals ..., thus decreasing the system capacity.\newline
3. Impacted modulation performance: ISI alters the frequency characteristics of the signals ..., affecting the transmission quality of the system.\newline
4. Increased signal power: In order to counteract the increased bit error rate caused by ISI ..., signal power and cost are increased.\newline
Therefore ..., appropriate modulation schemes, equalization techniques, and interference suppression methods are required to enhance system performance.
\end{CJK*}
		\\
		\bottomrule
	\end{tabularx}%
	\caption{Sample responses from three models.}
 \label{table4}%
\end{table*}%

\section{Conclusion and Future work}

In this paper, we introduce CourseGPT-zh, an open-source education large language model for professional courses, and the corresponding question-answering data construction framework. Based on this framework, we have implemented a communication principles specialized chatbot. Unlike previous work, we have focused on the comprehensiveness and diversity of questions, as well as the alignment of question responses with human needs. Furthermore, we have integrated discrete prompt optimization to enhance response quality, with LLM-as-judge for automatic multi-dimensional evaluation. Finally, we have trained CourseGPT-zh using the parameter-efficient fine-tuning method LoRA. Experimental results show that CourseGPT-zh exhibits impressive performance in question-answering for the corresponding course topics.

However, due to the limitations of the model's parameter size, particularly its pre-trained knowledge and reasoning capabilities, further work is needed to expand and enhance the performance of CourseGPT-zh. Future directions for development are outlined below.

\textbf{Reducing Hallucinations} \quad For specialized courses, the accuracy of responses is crucial for the user experience. However, even the most advanced models, such as GPT4, struggle with the issue of hallucinations, which is more pronounced in 6B parameter models. Furthermore, the performance of the 6B model in answering questions about specific professional knowledge is constrained by the limitations of its pre-trained corpus. To address these issues, we plan to construct a knowledge base of specialized knowledge based on textbooks and encyclopedias in the future and utilize Retrieval Augmented Generation (RAG) technology to provide references for response generation. This approach can reduce the model's hallucination problems and increase the accuracy of responses. Moreover, to address the potential changes in response distribution after incorporating references, we can use prompt optimization and joint adjustment with RAG to achieve the desired response style.

\textbf{High-quality Professional Knowledge Base} \quad CourseGPT-zh conducts knowledge extraction and question-answer pair construction based on a high-quality professional knowledge base. The richness and comprehensiveness of the professional knowledge base have a significant impact on the quality of the constructed data. Moreover, the most up-to-date knowledge base can prompt large models to generate the latest professional content. However, in the current work, data construction is only based on the main reference books of the corresponding courses. In the future, more field-related professional knowledge bases will be introduced to further improve the quality and quantity of the data and to further tap the potential of large language models.

\textbf{General Tasks and Extended Capabilities} \quad Currently, CourseGPT-zh is optimized for single-turn question-answering. However, multi-turn question-answering is also of great importance in future work. To improve the quality of multi-turn question-answering structured data in vertical domains, it is necessary to design a dedicated framework to ensure that large models can accurately understand contextual needs and refer to professional knowledge during the data construction process. Furthermore, other extended capabilities, such as Socratic teaching and problem-solving reasoning, are also very important for large language models in the field of education. These capabilities need to be further expanded in future work.

Finally, in light of the potential social risks posed by CourseGPT-zh, it is imperative to further enhance its security measures to prevent malicious utilization.





\bibliographystyle{unsrt}  
\bibliography{references}

\end{document}